# Uncertain and Approximative Knowledge Representation to Reasoning on Classification with a Fuzzy Networks Based System

**Mohamed Nazih Omri**

Département de Mathématiques et d'Informatique,
Institut Préparatoire aux études d'Ingénieur de Monastir,
Route de Kairouan, 5019 Monastir.
Tel. 216 3 500 273, Fax. 216 3 500 512
E-mail : Nazih.Omri@ipeim.rnu.tn

**Abstract**

The approach described here allows to use the fuzzy Object Based Representation of imprecise and uncertain knowledge. This representation has a great practical interest due to the possibility to realize reasoning on classification with a fuzzy semantic network based system. For instance, the distinction between necessary, possible and user classes allows to take into account exceptions that may appear on fuzzy knowledge-base and facilitates integration of user's Objects in the base. This approach describes the theoretical aspects of the architecture of the whole experimental A.I. system we built in order to provide effective on-line assistance to users of new technological systems: the understanding of "how it works" and "how to complete tasks" from queries in quite natural languages. In our model, procedural semantic networks are used to describe the knowledge of an "ideal" expert while fuzzy sets are used both to describe the approximative and uncertain knowledge of novice users in fuzzy semantic networks which intervene to match fuzzy labels of a query with categories from our "ideal" expert.

**1. Introduction**

Users do not learn through written instructions. Learning how to use a new technological system is mainly an exploratory activity. Exploring learning has shown to improve the abilities constructing to successful error handling and discovering and eventually constracting correct knowledge, but exploratory activity frequently leads to experience uninterested states or not reach the interested state goal. Users need assistance not only to avoid errors, but to understand how the system interprets their commands and How and Why to act in order to reach their goal system (Tijus, Poitrenaud, Richard & Leproux, in press).

In order to respond to a query, an executive assistant might know very precisely the goal the user has in mind, which means an object in a given state (the properties of the object being transformed). Moreover, even when goals are fairly well defined, it is often necessary to think about superordinate goals. Lets take (example1) the query of a subject using a Macintosh Computer.

**Example 1 :**

    **Subject :** "I can not understand why it can find Paint"
    (*Assistant's Understanding : To find [Paint])*
    **Assistant's :** "What is [Paint]?
    **Subject :** "Paint is a software to draw"
    **Assistant's diagnosis** : "as you are using a Text Editor you can only search for text files. You must return to the finder and search for the software. Only the finder gives access to softwares."But, why do you need to find Paint?
    **Subject :** Because I want to include a figure in the text.
    *Assistant's Understanding : To include [a figure, in the text]*
    **Assistant's :** "What is [a figure]?

This short verbal protocol shows that the assistant has mainly to understand goals and objects of the goals as well as superordinate goals. To diagnose the latter helps to provide guidance ( do you know that the text editor you are using makes possible to draw?).

## 2. The Semantic Net of the "Ideal Expert"

Executing a procedure serves to reach a Goal on an Object. The underlying psychological hypothesis is these Goals are Object properties, and as such, are generators of Object categories. Goals and procedures define the function of Objects and the way to use them. As functional properties of Objects, they enter into the construction of semantic networks in the same way as structural properties. We define a procedure as a sequence of operations whose execution serves to reach a Goal, and where the elements of the sequence are either primitive actions or subGoals which themselves call for associated procedures.
We define the "Ideal" Expert knowledge of a system as the knowledge that is sufficient to the system and that is described in a semantic network (figure 2).

Construction of the Ideal Expert Knowledge starts if given a set of Tasks that are executed using elements of one technical device through procedures. The first step is the task decomposition as a hierarchy of Goal decomposition into subGoals from the level of the Goal of the task to primitive actions. The second step consists in :
    **i -** drawing up a list of possible Goals and the procedures to reach these Goals,
    **ii -** constructing the Ideal Expert Net as a classical semantic network.
But, instead of using structural properties of system's interface Objects, Goals reachable with those Objects are used as properties. The ideal user's description uses valid procedures that have to be applied to the elements of the device in order to successfully complete the task. Classes of Objects and relations between classes of Objects merge from routines for classification and routines for classes organization. The above principle may be derived from an algebraic entity: the Galois lattice of a binary relation (Barbut and Monjardet, 1970). Given a set of objects and a set of properties, we can built the binary relation composed by the set of all couples object-property for which the object holds the property. Such a binary relation may be presented a in two way table with objects in lines and properties in columns, checking cells structural properties of geometrical shapes are organised, we obtain the binary relation shown by the (figure 2), called the Hass diagram of the lattice. Several algorithms have been proposed to built the Galois lattice of a binary relation (Norris, 1978; Ganter, 1984; Bordat, 1986). This kind of lattices, and the Hasse diagram which is used to draw them, have severed to investigate categorization problems (Guénoche & Van Mechelen, 1993; Storms, Van Mechelen & De Bocck, 1994). Figure (2) shows how the Galois lattice is used to formalize the class inclusion relation underlying data in table 1 by the mean of the double partial orders over:
    **i -** the set of objects,
    **ii -** the set of properties.
The resulting structure is a two way lattices in which, when a set of objects A includes a set of objects B, the set of properties of the objects in A are included into the set of properties of the objects in B, and vice versa.
The nodes of such a graph are called formal concepts by Wille (1982) because they clearly express the classical duality between the extension and the intension of concepts.

In summary, a basic Ideal Expert Net (figure 1) is a hierarchy of abstract and concrete classes represented as a single directed graph [G: (N, G, R)] which is defined to consist of a set of nodes N (representing classes of Objects), a set of attributes G (representing Goals) and a set of relations R (representing classes inclusion) between nodes.

Finally, the structural properties of Objects are added focusing on those that justify the application of the Procedures attached to the object in hand. Note that, by answering queries of the users while they try to perform a given goal, the Expert Assistant delivers not only planning information, but also a goal structure and the knowledge of what justifies the procedure by providing the knowledge that is included in the Ideal Expert Net.

If the Assistance System does not understand the meaning of an instruction, it discusses with the user until it is able to interpret the query in its own language. With the learning of new words in natural language as the interpretation produced in agreement with the user, the system improves its representation scheme at each experiment with a new user and, in addition, takes advantage of previous discussions with users :
    **i -** the standard Objects and recognized by the software are described in a semantic network where goals stand for properties of Objects,
    **ii -** as the queries of an user are expressed in natural language and as they correspond more or less to these standard denominations, the system establishes fuzzy connections between its primary knowledge and the new labels of Objects or procedures expressed by the user.

For instance, since a new attribute (new verb) can be identified with more than one primary attribute, we found convenient to use degrees of appropriateness of each such identification. For this purpose, we used the framework of fuzzy set theory and we describe fuzzy characterizations of classes of Objects with Goals as properties. Possibility and/or certainty measures enable us to quantify the opinion of users regarding the degree of similarity or convenience of associations between descriptions of both Objects and procedures.

In the next section, we present the proposed system dealing with degrees of membership of an Object or a procedure formulated in the user's language to one of the classes of the fuzzy semantic network. It takes into account the degree of inclusion between fuzzy classes and the grade of membership of an instance to one or several classes. Rossazza (90) introduced the concept of fuzzy Objects and Dubois and Prade (1989) addressed the problem of typicality of class hierarchy, but very few works have been proposed in this direction, specially as a way of planning actions.

## 3. The Fuzzy Semantic Net of Novice Users

However given the polysemic aspects of natural language (verbs and nouns which express goals and device objects), with the necessity of a man-machine interface that involve queries of users, the problem that is under investigation is how to match the content of a query (the label of an Object and the label of a Goal applied to this Object, as expressed by a novice user) to their corresponding items (class of Objects and Goals as properties) in the Ideal Expert Net.

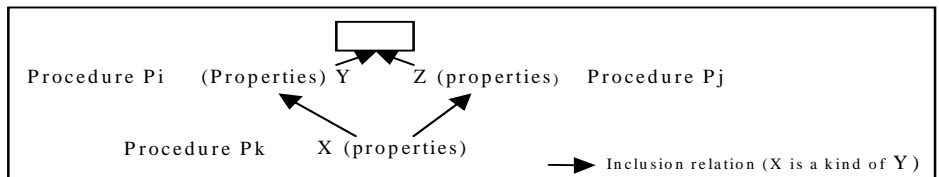

**Figure1:** *Procedural Semantic Net representation with inclusion relations. Procedural and declarative semantics of the device merges in regard of applied procedures. Classes Y and Z inherit of procedures of superordinate classes as class X inherits of procedures of both Y and Z classes (multiple inheritance).*

|  | Key | Forward-Word | Backward-Word | Forward-Char | Backward-Char | Char | Word | Unit | Direction |
|---|---|---|---|---|---|---|---|---|---|
| Direction (Forward) |  | X |  | X |  | X | X | X |  |
| Direction (Backward) |  |  | X |  | X | X | X | X |  |
| Choose |  | X | X | X | X | X | X |  | X |
| Select |  | X | X | X | X | X | X | X |  |
| Press | X |  |  |  |  |  |  |  |  |

**Table 1.**

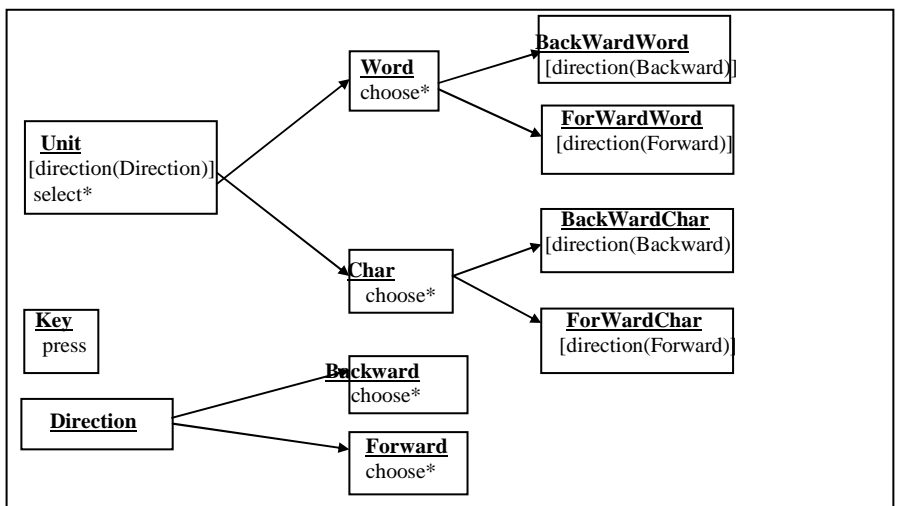

**Figure 2. Fuzzy Semantic Network**

The example of the technical system we consider here is a Word Processor software, with Objects such as "chain-of-characters", and procedures such as "cut" or "copy". For a novice user of the software, the list of standard denominations is not obvious and he often would like to ask an expert operator about how to execute an action such as "how to rub letters".

To describe common sense knowledge, different kinds of relationships must be used. They are necessary, sufficient and possible for Torasso and Console [Rossazza, 90], necessary and typical for Padgham [Padgham, 88]. At a theoretical level, the first organisation, as defined by Torasso and Console, appears to be powerful. However, it is difficult to define the sufficient linguistic value in practical applications and we do not use this concept [Tijus & Poitrenaud, 1992]. In our present work, we distinguish necessary properties from possible ones and we do not consider sufficient conditions. It follows that each Object or procedure is described by three fuzzy areas (necessary, possible and user areas), which describe the properties of the attributes A.

## 3.1. Fuzzy areas

**Necessary area :** The necessary area $Z^N$(P, A) of a linguistic value P of an attribute A is the set of couples (necessary linguistic value, necessity degree) admissible for A relatively to P. This area is fuzzy because some values are less admissible than others.

**Example 2 :**    **Attribute :** procedure
    **domain**:[EraseMenu,EraseWithkey, Select, CutWithMenu]
    **value**    :Erase
    **Necessary-area**:[(EraseWithMenu, 1) (EraseWithKey, 0.9) (CutWithMenu, 0.6)]

The value of $Z^N$(Erase, Procedure) is obtained from a necessity distribution. The latter is given by one expert using EraseWithMenu, EraseWithKey and CutWithMenu to design the system linguistic value Erase.

**Possible area:** The possible area $Z^P$(P, A) of a linguistic value P of an attribute A is the set of couples (possible linguistic value, possibility degree) admissible for A relatively to P. This area is also obtained from a possibility distribution given by a system expert.

**Example 3 :**    **attribute** :procedure
    **domain** :[EraseWithMenu, EraseWithKey, Select, CutWithMenu]
    **value**    :Erase
    **Possible-area**:[(EraseWithMenu, 1) (EraseWithKey, 1) (CutWithMenu, 0.8)]

**User area:** The user area $Z^U$(P, A) of a linguistic value P of an attribute A is the set of couples (user linguistic value, possibility degree) admissible for A relatively to P. This set is obtained from the user reasoning.

The necessary, the possible and the user areas allow to define attributes of Objects classes.

## 3.2. Fuzzy attributes

**a) System's fuzzy attributes :** The structure of System attributes A is as follows:

$$A:\left[\left(P_1\left(Z_1^N, Z_1^P\right)\right)...\left(P_k\left(Z_k^N, Z_k^P\right)\right)\right] \quad (1)$$

Where
$P_i$ is the i-th linguistic variable of A, $Z_i^N$ is the fuzzy necessary area associated with $P_i$, $Z_i^P$ is the fuzzy possible area associated with $P_i$ .

**Example 4 :**

**System goals** :[
    (**Erase**  ( **Necessary-area**:{(EraseWithMenu,1) (EraseWithKey, 0.9)  (CutWithMenu, 0.6)}
        **possible-area**:{(EraseWithMenu, 1) (EraseWithKey, 1) (CutWithMenu, 0.8)}))
    (**Select** ( **Necessary-area**:{(SelectToGoThrough, 1)(SelectTo Delimit, 0.8)}
        **possible-area**:{(SelectToGoThrough, 1)(SelectTo Delimit, 1)}))]

**b) User's fuzzy attributes** : The structure of the user's attribute has two kinds of descriptions. The first one is a set of fuzzy areas and the second one is a set of couples (linguistic possible value, possibility degree). They are shown as follows:

$$A:[A_1(Z_1^U)\ldots A_n(Z_n^U)]\qquad(2)$$

Where
   $A_i$ are user's linguistic variables of the attribute A,
   $Z_i^U$ the user's area fuzzy values.

For a couple (possible linguistic value, associate degree), this structure becomes:

$$A:[A_1\{(p_{11},d_{11}),\ldots,(p_{1k},d_{1k})\},\ldots,A_n\{(p_{n1},d_{n1}),\ldots,(p_{nl},d_{nl})\}]\qquad(3)$$

Where
   $A_i$ linguistic variables associated with A,
   $P_{ik}$ possible linguistic variables associated with $A_i$.
   $d_{ik}$ possible linguistic variables associated with $P_{ik}$.

The definition of the user's attribute is as follows:

**Definition 1:** Let $P_i$, i ∈ [1, n], a set of user's linguistic variables. We define a user's attribute $A^U$ by:

$$A^U = Z^U(P_1) \times Z^U(P_2) \times \ldots \times Z^U(P_n)\qquad(4)$$

**Example 5 :**

    **attribute**: procedure
    **domain** : [EraseWithMenu, EraseWithKey, Select, CutWithMenu]
    **value** : Gum
    **User's area**.: [(EraseWithMenu, 1) (EraseWithKey, 0.7) (CutWithMenu, 0.5)]

Three kinds of attributes can therefore be define for the object O: necessary attribute $A^N$, possible attribute $A^P$ and user attribute $A^U$. These attributes are fuzzy as results of the Cartesian products of fuzzy areas. The set of attributes describes classes and instances. Properties of $A^U$ are user properties which can't be used in the theoretical definition of classes and instances.

**Definition 2 :** Let $A_i$, i ∈ [1, n] be a set of attributes. We define class C as:

$$C = A_1 \times A_2 \times \ldots \times A_n\qquad(5)$$

**Definition 3 :** Let $a_i$, i ∈ [1, n] be a set of attributes $A_i$, i ∈ [1, n]. We define an instance I , with analogy to a class, as :

$$I = a_1 \times a_2 \times \ldots \times a_n\qquad(6)$$

## 4. Hierarchical relation

The integration of fuzzy properties; as 'to efface', in the Object's description implies the valuation of relations in [0, 1] (Omri, 1993; Omri 1994, Omri & Tijus 98). There are two kinds of relationships: the relation 'kind-of' between two classes and the relation 'is-a' between a class and an instance. One class may be a kind-of an other class, this to some extend. Each kind of relationship is described by a membership value obtained from the inclusion between areas or between attributes.

We define degrees of inclusion in the following sections for variables, attributes, classes and instances.

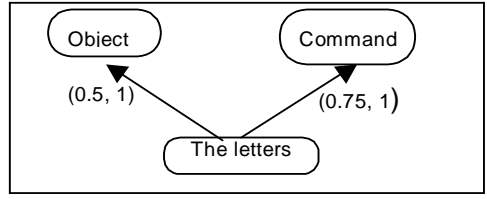
**Figure 3.a. :** valued "is-a" link

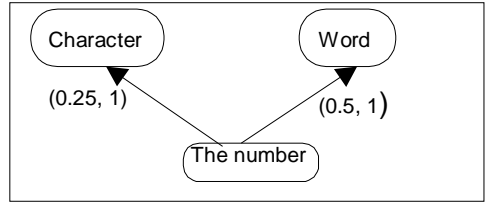
**Figure 3.b. :** valued "is-kind-of" link.

In figure 3.a, 'The letters' appears as a 'kind-of' 'Command' with the couple of values (0.75, 1), the first number is the necessity coefficient and the second is the possibility coefficient. It is also a 'kind-of' 'Object' with (.5, 1) as a couple of values. In figure 3.b, 'the number' is-a 'Character' with necessity coefficient 0.25, possibility coefficient 1 and is-a 'Word' with 0.5 and 1 as necessity and possibility coefficients respectively.

### 4.1. Inclusion between fuzzy linguistic variables

Let Y be a universe linguistic values, T and S are two linguistic variables defined on Y. $T = \{(t_1, d_1), \ldots, (t_n, d_{n1})\}$ and $S = \{(s_1, d'_1), \ldots, (s_m, d'_m)\}$, where $t_i$ and $s_j$, i ∈ [1, n], j ∈ [1, m], respectively the different linguistic values of T and S. $d_i$ and $d'_j$ degree associated with $t_i$ and $s_j$.

**Definition 4 :** Let $f^*_{T \cap S}$ be the membership function which results from the intersection of linguistic variables T and S defined on the same universe Y. We define inclusion degree of T in S by:

$$Deg^*(T \subset S) = \frac{\sum_{y \in Y} f^*_{T \cap S}(y)}{\sum_{y \in Y} f^*_T(y)} \qquad (7)$$

$f^*_T$ is the membership function of the linguistic variable T.
It can be applied for necessary and possible areas, * standing for N or P.

**a) Case of system attribute**

**Definition 5 :** Let $Deg^N(T \subset S)$ and $Deg^P(T \subset S)$ be the necessary and possible inclusion degrees of T in S. The resulting degree is obtained as follows:

$$Deg(T \subset S) = \frac{Deg^N(T \subset S) + Deg^P(T \subset S)}{2} \qquad (8)$$

In the particular case where T=S, we have $Deg(T \subset S) = 1$.

**b) Case of user's attribute**

**Definition 6 :** Let $f_T$ and $f_{T \cap S}$ be respectively the membership functions associated with T and $T \cap S$. We define the inclusion degree of T in S by:

$$Deg(T \subset S) = \frac{\sum_{y \in Y} f_{T \cap S}(y)}{\sum_{y \in Y} f_T(y)} \qquad (9)$$

**4.1.1. Inclusion between fuzzy attributes**

**Definition 7 :** Let $T_1, T_2, \ldots, T_i$, i linguistic variables for an attribute A, $S_1, S_2, \ldots, S_i$, i linguistic variables for an attribute B where $T_1 S_1, T_2 S_2, \ldots, T_i S_i$ respectively defined on the same universe. We define inclusion degree of A n B by:

$$Deg(A \subset B) = \min_{1 \leq p \leq i}\left(Deg(T_p \subset S_p)\right) \quad (10)$$

**4.1.2. Inclusion between fuzzy classes**

**Definition 8 :** Let $A_1, A_2, \ldots, A_n$ be n attributes which defines the fuzzy class $C_1$ and $B_1, B_2, \ldots, B_n$, n attributes which define the fuzzy class $C_2$. We define inclusion degree of $C_1$ in $C_2$ by :

$$Deg(C_1 \subset C_2) = \min_{1 \leq i \leq n}\left(Deg(A_i \subset B_i)\right) \quad (11)$$

**4.2. Instances and classes**

In the case of a class and instance, we deal with degrees of membership degrees. These degrees measure the physical representation of the class by the instances. They are obtained from inclusion degrees between fuzzy attributes. We define the membership of an instance I in class C by :

$$Deg(I \in C) = \min_{1 \leq i \leq n}\left(Deg(a_i \in A_i)\right) \quad (12)$$

**5. Conclusion**

Inspired by concepts produced in the framework of Object oriented programming tools, from researches in problem solving (Tauber, 1988) and from analysing tasks like text editing and the use of a sophisticated telephone (Richard J.F., Poitrenaud S. Tijus C., 1993), we implement the know-how (named "knowledge of an ideal expert") one needs to perfectly use any technical system with semantics networks focusing on classes of Objects and their categorisation via class inclusion from the point of view of the applied Goals they share: Goals are organized in a semantic network whose nodes are Object categories and whose arcs indicate the class inclusion relation (Poitrenaud, 94).